\documentclass[letterpaper, 10 pt, conference]{ieeeconf}  

\IEEEoverridecommandlockouts                              

\usepackage{graphicx} 
\usepackage{amsmath} 
\usepackage{amssymb}  
\usepackage[short,nocomma]{optidef}
\usepackage[dvipsnames]{xcolor}
\usepackage{stmaryrd}
\usepackage{algorithmic, algorithm}
\usepackage{tabularx}
\newcommand{\quotes}[1]{``#1''}

\title{\LARGE \bf
Shift-invariant waveform learning on epileptic ECoG
}

\author{Carlos H. Mendoza-Cardenas$^{1}$ and Austin J. Brockmeier$^{2}$
	\thanks{$^{1}$Carlos H. Mendoza-Cardenas is with the Department of Electrical \& Computer Engineering, University of Delaware, Newark, Delaware, USA. He has been partly funded by Minciencias (Colombia). {\tt\small cmendoza@udel.edu}}%
	\thanks{$^{2}$Austin J. Brockmeier is with the Department of Electrical \& Computer Engineering and the Department of Computer \& Information Sciences,  University of Delaware, Newark, Delaware, USA. {\tt\small ajbrock@udel.edu}}%
	\thanks{This work was generously supported by the Data Science Institute at the University of Delaware through the allocation of resources in the Caviness and DARWIN clusters.}
}

\usepackage{hyperref}
\begin{document}

This is the accepted version of the paper to appear in the 2021 43rd Annual International Conference of the IEEE Engineering in Medicine and Biology Society \href{https://embc.embs.org}{(EMBC)}.\\

\textsuperscript{\textcopyright} 2021 IEEE. Personal use of this material is permitted. Permission from IEEE must be obtained for all other uses, in any current or future media, including
reprinting/republishing this material for advertising or promotional purposes, creating new
collective works, for resale or redistribution to servers or lists, or reuse of any copyrighted
component of this work in other works.

\maketitle
\thispagestyle{empty}
\pagestyle{empty}

\begin{abstract}

Seizure detection algorithms must discriminate abnormal neuronal activity associated with a seizure from normal neural activity in a variety of conditions. Our approach is to seek spatiotemporal waveforms with distinct morphology in electrocorticographic (ECoG) recordings of epileptic patients that are indicative of a subsequent seizure (preictal) versus non-seizure segments (interictal). To find these waveforms we apply a shift-invariant k-means algorithm to segments of spatially filtered signals to learn codebooks of prototypical waveforms. The frequency of the cluster labels from the codebooks is then used to train a binary classifier that predicts the class (preictal or interictal) of a test ECoG segment. We use the Matthews correlation coefficient to evaluate the performance of the classifier and the quality of the codebooks. We found that our method finds recurrent non-sinusoidal waveforms that could be used to build interpretable features for seizure prediction and that are also physiologically meaningful.
\end{abstract}

\section{INTRODUCTION}

The unannounced occurrence of seizures is a major health risk factor in individuals with epilepsy. Since the abnormal neuronal activity associated with a seizure correlates with changes in the electroencephalographic (EEG) recording of a patient, the development of algorithms that can predict impending seizures from EEG data is an active research area \cite{Kuhlmann2018}. However, as was observed in a recent work \cite{Mendoza-Cardenas2021a}, most of the algorithms so far developed use EEG features that are computed without regard for the morphology of non-sinusoidal neuronal oscillations that pervade the EEG recording. In contrast, several works have shown the close connection between the waveform shape of neural oscillations and the physiology and pathophysiology of the brain \cite{Gerber2016,Cole2017,Cole2017a}, and especially its use in effectively discriminating between normal and abnormal oscillations in electrocorticographic (ECoG) recordings of epileptic patients \cite{Liu2018}.

Although several works have explored the use of shift-invariant and data-driven algorithms in EEG data \cite{Principe2015,Brockmeier2016,Jas2017,Dupre2018}, only one work have proposed a waveform learning method in the context of seizure prediction \cite{Cui2018}. In that work, the authors proposed a bag-of-waves (BoWav) representation, using a $k$-means clustering algorithm to learn two codebooks of interictal and preictal waveforms from a random sample of small, single-channel windows of a multi-channel EEG recording. Their codebooks are characterized by the presence of waveforms with similar shapes but with different time shifts. In contrast, we propose a shift-invariant method that mitigates this situation, which can increase the diversity of the morphologies represented by the codebooks.

Here, we explore a data-driven, shift-invariant waveform learning method that learns two class-specific sets (codebooks) of non-sinusoidal spatiotemporal patterns from the interictal and preictal states of epileptic ECoG recordings. We call those patterns prototypical waveforms, or centroids, as they approximate waveforms with similar morphology that occur repeatedly across time. The learning of the codebooks is performed through a shift-invariant $k$-means algorithm and optimized for the task of classifying between interictal and preictal ECoG segments. First, the multichannel ECoG data is spatially filtered with a pair of spatial filters that maximize the spectral band power of a signal from one condition while minimizing it for signals from the other condition. Second, the number of centroids (prototypical waveforms) in each codebook and the centroid length are selected through the cross-validation of a binary classifier that predicts the class (preictal vs. interictal) of an ECoG signal. Finally, the learning of the two codebooks is performed with the selected hyperparameters. In addition to the classification performance as an ensemble-level measure of the quality of the two codebooks, we ran a $\chi^2$ test on each prototypical waveform to quantify its predictive value.

\section{Methods}

\subsection{Data}

We used continuous long-term multichannel ECoG recordings from four epileptic patients publicly available at \verb|ieeg.org| \cite{Wagenaar2013}. Table~\ref{tab:data_info} presents some characteristics of the data: the patient's age in years; the number of channels ($C$) and seizures (N.S); and the seizure type (S.T): complex partial seizure (CPS) and generalized tonic-clonic seizure (GTC). We used the same seizure annotations and subset of EEG channels that were used in a previous work \cite{Kini2019}. The relevant clinical seizure markings for this work are the earliest EEG change (EEC), which is the point in time with the first clear and sustained change from the patient's EEG baseline before the seizure onset, and the end of the seizure. We assume that the activity happening between those two time points represents the ictal state. We define the interictal state as the activity that happens at least four hours away from the ictal state, and the preictal state as the activity that occurs in the 1-hour interval from 65 to 5 minutes before EEC, leaving five minutes of minimum intervention time.

\begin{table}[t]
	\caption{Data characteristics}
	\label{tab:data_info}
	\centering
	\begin{tabular}{lrrrl}
		\textbf{Name} & \textbf{Age} & \textbf{\textit{C}} & \textbf{N.S.} & \textbf{S.T.} \\ 
		\hline \\[-1.5ex]
		HUP111A  & 40 &  48  & 5    & CPS \\
		Study012 & 37 &  79  & 17  & CPS,GTC    \\
		Study017 & 39 &  16  & 9  & CPS,GTC    \\
		Study019 & 33 &  96  & 16  & CPS    \\
	\end{tabular}
\end{table}

\subsection{Preprocessing and spatial filtering}

We used the preprocessing and spatial filtering pipelines detailed in previous work \cite{Mendoza-Cardenas2021a}. The preprocessing removes the 60~Hz power line noise, discards some segments with artifacts like an additional broad-band peak at 60 Hz, and resamples the signal at 512~Hz. Let $\mathbf{X} \in \mathbb{R}^{M\times C\times L}$ be a $C$-channel ECoG segment split into $M$ non-overlapping windows of length $L$. Let $g_\mathbf{w}: \mathbb{R}^{M\times C\times L} \rightarrow \mathbb{R}^{M \times L}$ denote the spatial filtering step performed with the filter $\mathbf{w} \in \mathbb{R}^C$, such that $\check{\mathbf{X}} = g_\mathbf{w}(\mathbf{X})$ is a spatially-filtered ECoG segment that results from left multiplying the last two dimensions of $\mathbf{X}$ by $\mathbf{w}^\text{T}$. We used the common spatial patterns (CSP) method \cite{Blankertz2008} to compute two spatial filters, $\mathbf{w}_{0,\text{B}}$ and $\mathbf{w}_{1,\text{B}}$, such that they maximize the energy of band-passed filtered signals from the interictal and preictal state, respectively, while minimizing it for signals from the other state. We computed $\mathbf{w}_{0,\text{B}}$ and $\mathbf{w}_{1,\text{B}}$ for each of the nine passbands $\text{B} \in \{\delta[1.5-4], \theta[4-8], \alpha[8-15], \beta_{low}[15-26], \beta_{high}[26-35], \gamma_{low}[35-50], \gamma_{mid}[50-74], \gamma_{high}[76-120], \mathrm{HFO}[120-220]\}$, in units of Hz. To quantify the performance of the CSP filters in discriminating between the interictal and preictal conditions, a test set consisting of 1-minute ECoG segments from both conditions is passed as input to both CSP filters, and the energy of their output is compared to a threshold by a binary classifier. For this work, we chose the spectral band where that binary classifier had the highest area under the precision-recall curve. For simplicity, we will drop the subscript B in $\mathbf{w}_{\cdot,\text{B}}$ henceforth.

\subsection{Spherical shift-invariant k-means}

Here we describe the shift-invariant clustering algorithm that we use to explain a set of spatially-filtered ECoG signals using a scaled and shifted combination of prototypical waveforms forming a codebook. $\mathcal{X} = \{\mathbf{x}_n \in \mathbb{R}^L\}^{N}_{n=1}$ is a set of $N$ signals of length $L$, $T_P(\mathbf{x},\tau) = [x_{\tau},\,x_{\tau+1},\,\ldots,\,x_{\tau+P-1}]$ is a $P$-length window extracted from $\mathbf{x}$ at time shift $\tau$, with $P \leq L$ and \mbox{$\tau \in \{0,\ldots,L-P\}$}, $\mathbf{C} = [\mathbf{c}_1,\, \mathbf{c}_2,\,\ldots,\,\mathbf{c}_k] \in \mathbb{R}^{P\times k}$ is a matrix whose columns are a set of $k$ centroids of length $P$ that approximate the repeated occurence of waveforms in $\mathcal{X}$ at different time points, and $\nu \in [k]$ is the centroid index closest to $\mathbf{x}$, with $[k] = \{1,\ldots,k\}$. We use a spherical shift-invariant $k$-means algorithm to find the codebook $\mathbf{C}$ and the set $\{(\nu_n,\tau_n)\}^N_{n=1}$ that minimize the sum of distances from the signals in $\mathcal{X}$ to their closest centroid. This can be cast as the optimization problem

\begin{mini}
	{\substack{\mathbf{C}\\\{\nu_n,\tau_n\}^N_{n=1}}}
	{\sum_{n=1}^{N}d(T_P(\mathbf{x}_n,\tau_n),\mathbf{C}\mathbf{e}_{\nu_n}),}
	{\label{eq:sikmeans_optim}}{}
\end{mini}

\noindent with $\mathbf{e}_i$ being the standard unit vector in $\mathbb{R}^k$, with 1 in the $i$-th element and 0 everywhere else, such that $\mathbf{Ce}_i = \mathbf{c}_i$, and
\begin{equation}
d(\mathbf{y},\mathbf{z}) = 1 - \frac{\mathbf{y}^\text{T}\mathbf{z}}{\lVert \mathbf{y} \rVert \lVert \mathbf{z} \rVert}
\end{equation}

\noindent being the cosine distance between a pair of vectors $\mathbf{y}$ and $\mathbf{z}$ in $\mathbb{R}^P$. We use cosine instead of Euclidean distance because it is invariant to scaling by a non-negative value.

As an extension of the classic k-means algorithm, the sum in Eq.~\eqref{eq:sikmeans_optim} is minimized in two alternating steps. First, with a fixed codebook $\mathbf{C}$, the algorithm has a cluster assignment step where, for each signal $\mathbf{x}_n \in \mathcal{X}$, we find its closest centroid $\mathbf{c}_{\nu_n^*}$ and the best time shift $\tau_n^*$

\begin{argmini}
	{\nu_n,\tau_n}
	{d(T_P(\mathbf{x}_n,\tau_n),\mathbf{c}_{\nu_n}),\quad n\in [N].}
	{\label{eq:assignment_step}}{\nu_n^*, \tau_n^* =}
\end{argmini}

\noindent The set of tuples $\{(\nu^*_n, \tau^*_n)\}_n$ where $\nu^*_n = j$ defines the $j$-th cluster of signals in $\mathcal{X}$ whose closest centroid is $\mathbf{c}_j$: $\mathcal{C}_j = \{\mathbf{x}_n \in \mathcal{X} \mid \nu^*_n = j \}$, for $j \in [k]$.

Second, with the cluster assignments and time shifts fixed, we update the centroids

\begin{equation}
\mathbf{c}_j = \frac{1}{\vert \mathcal{C}_j\vert} \sum_{n \in [N]: \mathbf{x}_n \in \mathcal{C}_j} T_P(\mathbf{x}_n,\tau^*_n), \quad j \in [k],
\label{eq:update_step}
\end{equation}


\noindent with $\lvert \cdot \rvert$ denoting set cardinality. The centroids are initialized by picking $k$ signals from $\mathcal{X}$ at random and choosing the $P$-length window with the highest energy on each signal. A more involved initialization such k-means++, which ensures diversity of the initial centroids, could also be used.

To learn two class-specific codebooks, $\mathbf{C}_0 \in \mathbb{R}^{P \times k}$ for interictal and $\mathbf{C}_1\in \mathbb{R}^{P \times k}$ for preictal, we first spatially filter the interictal (preictal) data with $\mathbf{w}_0$ ($\mathbf{w}_1$), the CSP filter that maximizes the spectral band power for that specific class. Then, we use the shift-invariant $k$-means algorithm described here to learn $\mathbf{C}_0 \in \mathbb{R}^{P \times k}$ and $\mathbf{C}_1\in \mathbb{R}^{P \times k}$ from the spatially-filtered interictal and preictal data, respectively.

\subsection{Waveform counts}

We now explain how we use $\mathbf{C}_0$ and $\mathbf{C}_1$ to characterize a spatially-filtered ECoG segment in terms of waveform counts: the number of times a prototypical waveform from $\mathbf{C}_0$ or $\mathbf{C}_1$ \quotes{occurs} in that segment. We say that a waveform \quotes{occurs} in a segment if it is the closest centroid for a given window from that segment. Let $\mathbf{X} \in \mathbb{R}^{M \times C \times L}$ be a $C$-channel ECoG segment split into $M$ windows of length $L$ and with true class label $y \in \{0,1\}$, 0 for interictal and 1 for preictal. $\mathbf{X}$ is spatially filtered with $\mathbf{w}_0$ and $\mathbf{w}_1$: $g_{\mathbf{w}_0}(\mathbf{X}) = \check{\mathbf{X}}_0 \in \mathbb{R}^{M\times L}$ and $g_{\mathbf{w}_1}(\mathbf{X}) = \check{\mathbf{X}}_1 \in \mathbb{R}^{M\times L}$. Let $\texttt{assign}:~ \mathbb{R}^{M\times L}\times\mathbb{R}^{P\times k} \rightarrow [k]^M$ be a cluster assignment map (see Eq. \eqref{eq:assignment_step}) such that $\boldsymbol{\nu} = \texttt{assign}(\check{\mathbf{X}}, \mathbf{C}) = [\nu_1, \ldots, \nu_M]$, with $\nu_i \in [k]$ being the label of the prototypical waveform in $\mathbf{C}$ closest to the $i$-th window in $\check{\mathbf{X}}$, for $i \in [M]$. $\check{\mathbf{X}}_0$ and $\check{\mathbf{X}}_1$ are clustered using $\mathbf{C}_0$ and $\mathbf{C}_1$, respectively: $\boldsymbol{\nu}_0 = \texttt{assign}(\check{\mathbf{X}}_0, \mathbf{C}_0)$ and $\boldsymbol{\nu}_1 = \texttt{assign}(\check{\mathbf{X}}_1, \mathbf{C}_1)$.


Let $\texttt{BoW}: [k]^M \rightarrow \{0,\ldots,M\}^k$ be a bag-of-words mapping such that $\mathbf{z} = \texttt{BoW}(\boldsymbol{\nu}) = [z_1,\ldots,z_{k}]$ are the counts of the labels in $\boldsymbol{\nu}$, with $z_i = {\sum_j \llbracket \nu_j = i \rrbracket}$, for $i \in [k]$ and $j \in [M]$, and $\llbracket \textit{q} \rrbracket$ equal to 1 if \textit{q} is true, or 0 otherwise. We can then compose and concatenate all these steps to build a feature vector $\breve{\mathbf{z}} \in \mathbb{R}^{2k}$ from the ECoG segment $\mathbf{X}$ as

\begin{equation}
\begin{split}
\breve{\mathbf{z}} = [&\texttt{BoW}(\texttt{assign}(g_{\mathbf{w}_0}(\mathbf{X}), \mathbf{C}_0)),\\
              &\texttt{BoW}( \texttt{assign}(g_{\mathbf{w}_1}(\mathbf{X}), \mathbf{C}_1))]
\end{split}
\label{eq:BoW}
\end{equation}

\subsection{Classifier}

Let $h:\{0,\ldots,M\}^{2k}\rightarrow \{0,1\}$ be a binary classifier such that $\hat{y} = h(\breve{\mathbf{z}})$ is the class label estimate of an ECoG segment $\mathbf{X} \in \mathbb{R}^{M \times C \times L}$ with a feature vector $\breve{\mathbf{z}}$ computed as in Eq. \eqref{eq:BoW}. We consider a multinomial naive Bayes (\texttt{MultinomialNB}) classifier and a logistic regression (\texttt{LogisticRegression}) classifier. \texttt{MultinomialNB} uses the waveform counts and training labels to estimate the prior class (interictal/preictal) probabilities and the likelihood of the prototypical waveforms in $\mathbf{C}_0$ and $\mathbf{C}_1$ for each segment class; $\hat{y}$ is the maximum a posteriori (MAP) estimate. \texttt{LogisticRegression} learns a weight vector $\boldsymbol{\theta} \in \mathbb{R}^{2k}$ that linearly combines the features to maximize the log-likelihood of $Y$ (the class label); $\hat{y} = h(\breve{\mathbf{z}}) = \llbracket \sigma(\boldsymbol{\theta}^\text{T}\breve{\mathbf{z}}+\eta) > 0.5 \rrbracket$, with $\sigma(x) = 1/(1+ e^{-x})$, and $\sigma(\boldsymbol{\theta}^\text{T}\breve{\mathbf{z}}+\eta)=\mathbb{P}(Y=1 \mid X; \boldsymbol{\theta})$, the probability that $Y=1$, given the data $X$.

For \texttt{LogisticRegression}, we fit the intercept $\eta$ and use the $l_2$-norm penalty term with regularization parameter $\texttt{C}$. Furthermore, to downweight overly frequent waveforms, we use a feature scaling from the information retrieval literature called \textit{inverse \quotes{document}-frequency scaling} (idf), where \quotes{document} here means an ECoG segment. In particular, we used $\text{idf}(i) = \log((1+n)/(1+\text{df}(i)))+1$, where $n$ is the number of segments and $\text{df}(i)$ is the number of segments where the $i$-th prototypical waveform occurs.

\subsection{Waveform ranking} \label{subsec:waveform_ranking}

Let $\breve{\mathbf{C}} = [\mathbf{C}_0, \, \mathbf{C}_1] \in \mathbb{R}^{P\times2k}$ be a master codebook that results from concatenating the learned interictal and preictal codebooks, $\breve{\mathbf{c}}_i$ be the $i$-th waveform in $\breve{\mathbf{C}}$, $\mathrm{O}_{i,y}$ be the number of observed occurrences of $\breve{\mathbf{c}}_i$ in class $y \in \{0,1\}$, and $M_0$ and $M_1$ be the number of interictal and preictal windows in the test set, respectively.  We then build a contingency table~(see Table~\ref{tab:contigency_table}) for each waveform in $\breve{\mathbf{C}}$ and compute the $\chi^2$ test statistic to rank the prototypical waveforms according to the strength of the association between their occurrence and the window class.\\

\begin{table}[b]
	\caption{General form of a contingency table}
	\label{tab:contigency_table}
	\centering
	\begin{tabular}{l|rr}
		& Interictal & Preictal \\
        \hline
		$\breve{\mathbf{c}}_i$ occurs & $\mathrm{O}_{i,0}$ & $\mathrm{O}_{i,1}$ \\
		$\breve{\mathbf{c}}_i$ does not occur & $M_0 - \mathrm{O}_{i,0}$ & $M_1 - \mathrm{O}_{i,1}$
	\end{tabular}
\end{table}

\section{Results}

We took at random 763 interictal segments and 229 preictal segments from the ECoG recordings of each patient. A segment has $M=60$ non-overlapping windows 1-second long ($L=512$). We did an 80/20 split to create the training and test sets. The training set is used to learn the interictal and preictal codebooks, and to train the classifier $h$.

We used a 10-fold cross-validation to select the regularization parameter \texttt{C} for \texttt{LogisticRegression}, the number of centroids $k \in \{4, 8, 16, 32, 64, 128\}$ and the centroid length $P \in \{30, 40, 60, 120, 200, 350\}$ that yielded the highest average Matthews correlation coefficient (MCC)~\cite{Chicco2020}. Recall that the spectral band for the CSP filters was chosen by an early-stage energy-based classifier.

After training the classifier and learning the interictal and preictal codebooks using the hyperparameters chosen during cross-validation, we quantified the discriminatory power of the learned codebooks using the test set. Table~\ref{tab:classification_scores} shows the test MCC, precision, and recall scores of the two classifiers. Table~\ref{tab:best_hyperparameters} shows the spectral band chosen from the spatial filtering pipeline and the hyperparameters chosen during cross-validation.

\begin{table}[b]
	\caption{Classification scores}
	\label{tab:classification_scores}
	\centering
	\begin{tabularx}{0.95\columnwidth}{lc@{\hskip.2cm}c@{\hskip.2cm}c|c@{\hskip.2cm}c@{\hskip.2cm}c}
		& \multicolumn{3}{c|}{\texttt{LogisticRegression}} & \multicolumn{3}{c}{\texttt{MultinomialNB}}\\
		& MCC & Precision & Recall & MCC & Precision & Recall\\
		\hline
		Study019  & 0.841  &  0.927  & 0.826  & 0.718 & 0.935 & 0.630 \\
		HUP111A   & 0.489  &  0.541  & 0.717  & 0.452 & 0.535 & 0.652 \\
		Study012  & 0.400  &  0.429  & 0.783  & 0.327 & 0.420 & 0.630 \\
		Study017  & 0.262  &  0.373  & 0.609  & 0.248 & 0.400 & 0.478		
	\end{tabularx}
\end{table}

\begin{table}[b]
	\caption{Best hyperparameters}
	\label{tab:best_hyperparameters}
	\centering
	\begin{tabularx}{\columnwidth}{lrrrr|rr}
		& & \multicolumn{3}{c|}{\texttt{LogisticRegression}} & \multicolumn{2}{c}{\texttt{MultinomialNB}}\\
		& Band & \texttt{C} & $k$ & $P$ & $k$ & $P$\\
		\hline
		Study019 & $\gamma_{high}$ & 1   &  128 & 350 & 128 & 30 \\
		HUP111A  & $\theta$ & 1   &  128 &  30 &  64 & 30 \\
		Study012 & $\gamma_{high}$ & 2   &   32 & 120 & 128 & 60 \\
		Study017 & $\gamma_{low}$ & 0.5 &  128 & 120 & 128 & 120\\
		\hline		
	\end{tabularx}
	\footnotesize \texttt{C}: Regularization parameter, $k$: Number of centroids, $P$: Centroid length

\end{table}

\begin{figure}[hbt]
	\includegraphics[width=\columnwidth]{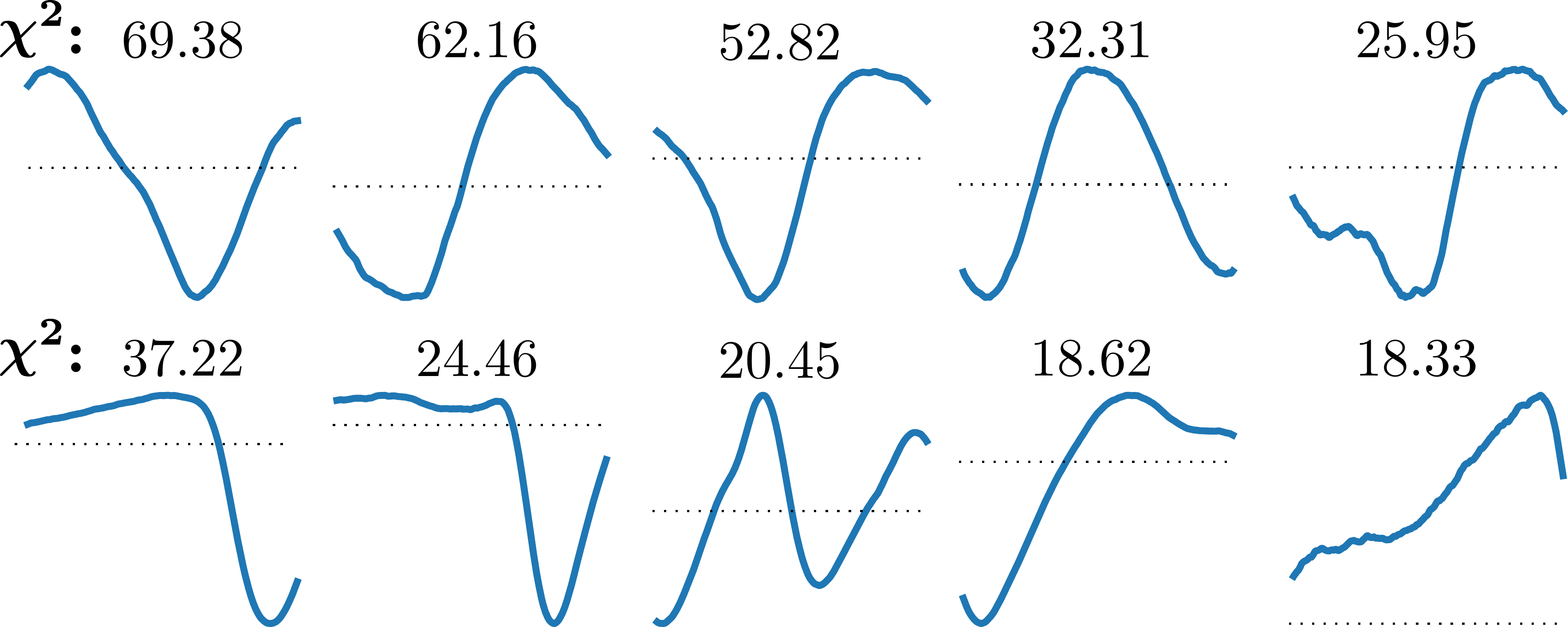}
	\caption{Top 5 waveforms on each class-specific codebook of Study019, ranked by their $\chi^2$-test value in the master codebook. Top row: preictal. Bottom: Interictal. Dotted line indicates zero amplitude.}
	\label{fig:top5_centroids}
	
\end{figure}

The performance of the classifier can be interpreted as a global measure of how well each (preictal or interictal) codebook represents the state it was trained for, and at the same time how different are those representations from each other. To assess and rank the predictive value of each individual learned prototypical waveform, we concatenated the codebooks learned for Study019, with $k=128$ and $P=~350$ (684 ms), into a single master codebook, as described in section \ref{subsec:waveform_ranking}, and applied a $\chi^2$ test on a contingency table for each waveform. Table~\ref{tab:example_contigency_table} shows the contingency table of the prototypical waveform with the highest $\chi^2$ (69.38) in the master codebook. Fig.~\ref{fig:top5_centroids} shows the top 5 preictal and interictal centroids with the highest $\chi^2$ score.

\begin{table}[t]
	\caption{Contingency Table~example}
	\label{tab:example_contigency_table}
	\centering
	\begin{tabular}{l|rr}
		& Interictal & Preictal \\
		\hline
		$\breve{\mathbf{c}}_{133}$ occurs & 19 & 41 \\
		$\breve{\mathbf{c}}_{133}$ does not occur & 9,161 & 2,719
	\end{tabular}
\end{table}

\begin{figure}[t]	
	\includegraphics[width=\columnwidth]{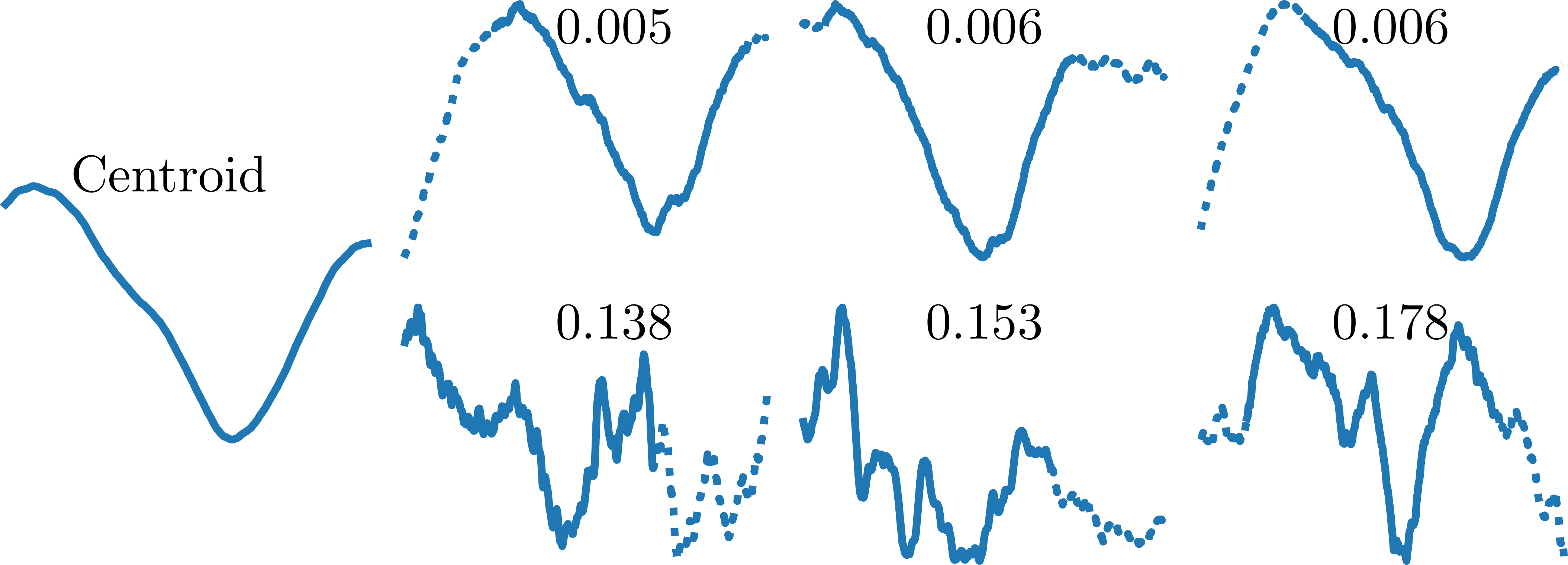}
	\caption{Preictal centroid with highest $\chi^2$ value in Study019 and the three closest and farthest test windows assigned to its cluster. On top of each test window, the cosine distance to the centroid. For the test windows, the solid line highlights the best time shift.}
	\label{fig:example_clusters}
\end{figure}

Fig. \ref{fig:example_clusters} shows a sample of the test windows assigned to a centroid when performing the step $\texttt{assign}(g_{\mathbf{w}_1}(\mathbf{X}), \mathbf{C}_1)$, and illustrates the shift-invariance property of our method.
	
\section{Conclusion}

Our shift-invariant waveform learning method was able to learn two class-specific codebooks of prototypical waveforms from the interictal and preictal state of epileptic electrocorticographic (ECoG) recordings that exhibited a high discriminative performance for one of four patients (Study019), as measured by the Matthews correlation coefficient of a binary classifier. The performance for the other patients varies from medium to low. A more detailed inspection of the data and the codebooks learned is required to understand the reason behind this significant variability between subjects in the classification performance. Notice, however, that our method uses only one CSP filter per class-specific codebook and that we do not perform any type of temporal filtering. We hypothesize that the addition of more spatial filters and the temporal filtering of the data could improve the discrimination between classes and the discovery of long-duration high-frequency oscillations that might be masked by the typical 1/$f$ spectrum of the ECoG signal. We also highlight that in addition to the good predictive power exhibited on some subjects, our method delivers waveforms with diverse morphologies and that can have a meaningful physiological interpretation. There are several directions for future work that can expand and improve the method here proposed. First, the statistical analysis of the spatial and temporal (relative to seizure onset) location of the waveforms clustered around each centroid could provide important insights not only for the development of a seizure prediction algorithm but also for our understanding of the pathophysiology and etiology of seizures. Second, automatic methods for the characterization of waveform shape and envelope could also be developed and applied to the learned waveforms. An example of this is the coefficient of variation of the envelope to quantify the rhythmicity of EEG waveforms \cite{Diaz2018}. Finally, more advanced methods for waveform learning, like convolutional dictionary learning \cite{Brockmeier2016,Jas2017,Dupre2018}, could also be explored in the context of seizure prediction.



\bibliographystyle{IEEEtran}	
\bibliography{EMBC21,NER21}

\end{document}